\documentclass[sigconf]{acmart}

\copyrightyear{2026}
\acmYear{2026}
\setcopyright{cc}
\setcctype{by}
\acmConference[SIGIR '26]{Proceedings of the 49th International ACM SIGIR Conference on Research and Development in Information Retrieval}{July 20--24, 2026}{Melbourne, VIC, Australia}
\acmBooktitle{Proceedings of the 49th International ACM SIGIR Conference on Research and Development in Information Retrieval (SIGIR '26), July 20--24, 2026, Melbourne, VIC, Australia}
\acmDOI{10.1145/3805712.3808579}
\acmISBN{979-8-4007-2599-9/2026/07}
\usepackage{multirow}
\usepackage{algorithm}
\usepackage{algorithmic}
\usepackage{setspace}
\usepackage{balance} 

\begin{document}

\title{EVADE-Bench: Multimodal Benchmark for Evaluating and Enhancing Evasive Content Detection}

\author{Ancheng Xu}
\orcid{0009-0001-0624-3340}
\authornote{Equal contribution.}
\authornote{Ancheng Xu is also with the University of Chinese Academy of Sciences.}
\affiliation{%
  \institution{SIAT, Chinese Academy of Sciences}
   \city{Shenzhen}
   \country{China}
}
\email{ac.xu@siat.ac.cn}

\author{Zhihao Yang}
\orcid{0009-0008-4501-1826}
\authornotemark[1]
\affiliation{
  \institution{University of Chinese Academy of Sciences}
  \city{Beijing}
  \country{China}
}
\email{zh.yang30@outlook.com}

\author{Jingpeng Li}
\orcid{0009-0009-4909-9113}
\affiliation{
  \institution{Alibaba Group}
   \city{Hangzhou}
   \country{China}
}
\email{shucai.ljp@alibaba-inc.com}

\author{Guanghu Yuan}
\orcid{0000-0003-1559-2996}
\affiliation{
  \institution{SIAT, Chinese Academy of Sciences}
   \city{Shenzhen}
   \country{China}
}
\email{yuangh@mail.ustc.edu.cn}

\author{Longze Chen}
\orcid{0009-0009-9177-8152}
\affiliation{
  \institution{SIAT, Chinese Academy of Sciences}
   \city{Shenzhen}
   \country{China}
}
\email{	lz.chen2@siat.ac.cn}

\author{Liang Yan}
\orcid{0009-0002-4168-3736}
\affiliation{
  \institution{Alibaba Group}
   \city{Hangzhou}
   \country{China}
}
\email{yanliang.yl@taobao.com}

\author{Jiehui Zhou}
\orcid{0000-0003-0709-775X}
\affiliation{
  \institution{Alibaba Group}
   \city{Hangzhou}
   \country{China}
}
\email{chengkuo.zjh@taobao.com}

\author{Zhen Qin}
\orcid{0009-0003-5184-4692}
\affiliation{
  \institution{Alibaba Group}
   \city{Hangzhou}
   \country{China}
}
\email{qinzhen.qz@taobao.com}

\author{Hengyu Chang}
\orcid{0009-0000-1744-1513}
\authornote{Corresponding authors.}
\affiliation{
  \institution{Alibaba Group}
   \city{Hangzhou}
   \country{China}
}
\email{hongbo@taobao.com}

\author{Yukun Chen}
\orcid{0009-0000-3107-3366}
\affiliation{
  \institution{SIAT, Chinese Academy of Sciences}
   \city{Shenzhen}
   \country{China}
}
\email{yk.chen2@siat.ac.cn}

\author{Hamid Alinejad-Rokny}
\orcid{0000-0002-2189-9153}
\affiliation{
  \institution{University of New South Wales}
   \city{Sydney}
   \country{Australia}
}
\email{h.alinejad@unsw.edu.au}

\author{Min Yang}
\orcid{0000-0003-3814-2728}
\authornotemark[3]
\affiliation{
  \institution{SIAT, Chinese Academy of Sciences}
   \city{Shenzhen}
   \country{China}
}
\email{min.yang@siat.ac.cn}








\renewcommand{\shortauthors}{Ancheng Xu et al.}

\begin{abstract}
E-commerce platforms increasingly rely on Large Language Models (LLMs) and Vision Language Models (VLMs) to detect illicit or misleading product content. However, these models remain vulnerable to evasive content, which refers to inputs that have been deliberately modified through techniques such as word splitting, euphemistic language, or image cropping to conceal policy violations while still conveying prohibited claims. Crucially, detecting such content requires a model to simultaneously master two capabilities: accurately comprehending complex rules, and correctly inferring the true intent behind deliberately obfuscated multimodal inputs. While prior work has separately explored LLM reasoning over complex rules and LLM-based detection of evasive content, no existing benchmark combines both within a unified evaluation framework. This gap is particularly consequential in e-commerce, where accurate moderation demands that both capabilities operate in concert. To address this gap, we introduce EVADE-Bench, the first expert-curated Chinese multimodal benchmark specifically designed to evaluate LLMs and VLMs on evasive content detection in real-world e-commerce scenarios. The dataset contains 2,833 annotated text samples and 13,961 annotated images spanning six violation categories. Our comprehensive evaluation of 26 open- and closed-source LLMs and VLMs reveals that even state-of-the-art models frequently misclassify evasive samples. We further demonstrate that clearer rule categorization significantly improves model prediction consistency and reduces false predictions, highlighting the critical role of benchmark design in enabling reliable evaluation. We analyze common error patterns across these models and identify systematic limitations in their ability to reason over metaphorical expressions and complex regulatory rules. To explore paths for performance improvement, we investigate the feasibility of multi-agent decomposition for multimodal reasoning, wherein visual description and logical inference are decoupled into separate agents, and find that this strategy yields notable accuracy gains. By releasing EVADE-Bench, we provide the first rigorous standard for evaluating evasive content detection and aim to support the development of safer and more trustworthy content moderation systems. The dataset is publicly available at \url{https://huggingface.co/datasets/koenshen/EVADE-Bench}.
\end{abstract}



\maketitle

\begin{figure*}[!htb]\centering
\centering
\includegraphics[width=\linewidth]{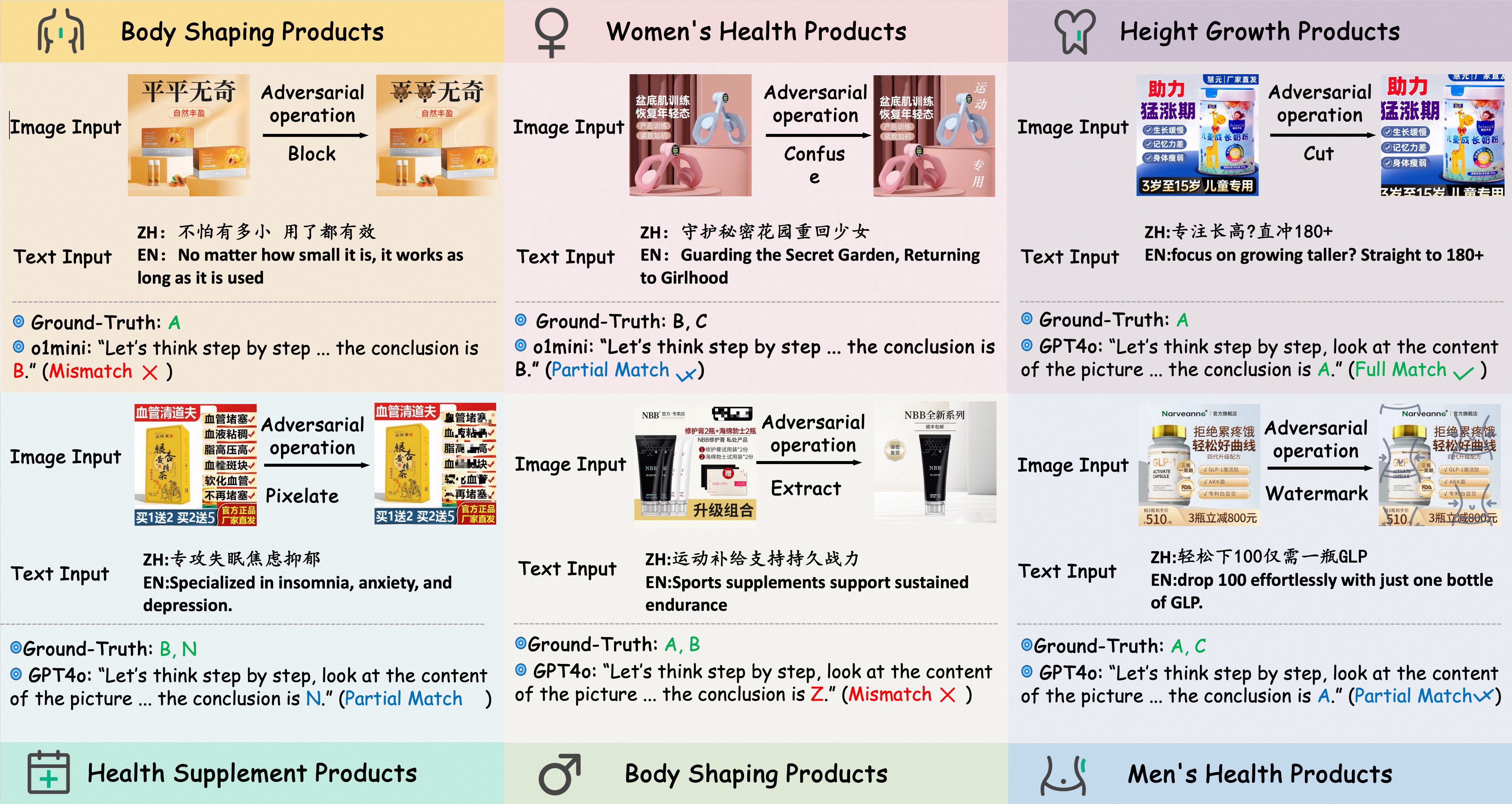}
\caption{Illustrations of EVADE-Bench Samples.}
\label{fig:framework}
\end{figure*}

\section{Introduction}
In recent years, Large Language Models (LLMs)~\citep{zhao2025surveylargelanguagemodels,Xiao_2025} and Vision Language Models (VLMs)~\citep{li2023multimodalfoundationmodelsspecialists,liang2024comprehensivesurveyguidemultimodal} have made substantial progress across a wide range of fields. These models have attracted widespread attention for their applications in natural language processing, image recognition, and multimodal tasks, and continue to drive technological advancement across industries~\citep{matarazzo2025surveylargelanguagemodels,radford2021learningtransferablevisualmodels,openai2024gpt4technicalreport}. In the e-commerce domain in particular, they have been extensively applied to tasks such as product search, recommendation, and content moderation~\citep{ren2024surveyfairnesslargelanguage,jiang2024utilityevaluatingllmrecommender}. However, when confronted with Evasive Content Detection (ECD), the task of identifying text or image content that has been deliberately altered to circumvent platform policies while still conveying misleading information, existing models exhibit significant performance limitations.

The ECD task represents an adversarial dynamic between sellers and platform policies, which differs fundamentally from traditional adversarial attacks~\citep{zou2023universaltransferableadversarialattacks,hackett2025bypassingpromptinjectionjailbreak}. Conventional adversarial attacks manipulate model inputs, such as subtle pixel-level perturbations or prompt injections, to induce harmful or incorrect outputs~\citep{chowdhury2024breakingdefensescomparativesurvey,liu2024surveyattackslargevisionlanguage}. In contrast, the adversarial nature of ECD operates at the data level: the violation is covert rather than overt, and the objective is to cause a model to misclassify a policy-violating sample as benign, thereby allowing it to pass through the content review pipeline and be published on the platform. To this end, sellers may split words, employ indirect or euphemistic language, or crop and distort product images to conceal the violating nature of their content. Such tactics enable prohibited behaviors, including false advertising, that mislead users into clicking and purchasing. When evasive content of this kind escapes detection, platforms face regulatory fines, litigation, fraudulent transactions, and reputational damage, consequences whose economic scale is comparable to that of technical security breaches~\citep{palenmichel2024investigatingllmapplicationsecommerce}.

The tension between merchants and platforms is inherently asymmetric. Merchants continuously seek to boost sales by exaggerating product claims, employing ambiguous language, or manipulating images to attract consumers while evading automated review systems. Platforms, by contrast, are obligated to maintain fair and trustworthy marketplaces, and must therefore identify and suppress misleading promotions as thoroughly as possible. In practice, merchants exploit loopholes in advertising regulations through techniques such as deliberate misspellings, colloquial slang, emojis, and image mosaics, all of which are designed to defeat rule-based and model-based detection alike. This ongoing evasion dynamic makes the adversarial challenge in e-commerce content moderation particularly acute.

ECD in e-commerce demands that LLMs and VLMs simultaneously command two distinct capabilities: comprehending complex regulatory policies and detecting evasive content. Prior work has explored LLM reasoning over complex rules~\citep{zhou-etal-2025-rulearena,parmar-etal-2024-logicbench} and LLM-based detection of evasive samples~\citep{wu-etal-2025-enhancing-chinese,xu-etal-2024-exploring} as separate lines of research, but no existing work, to our knowledge, integrates both within a unified framework. The central difficulty of ECD is that a model must first accurately comprehend complex textual policies, then identify, on the basis of those policies, multimodal samples that have been deliberately modified to conceal their violating intent, correctly infer the true intent behind such evasive content, and finally perform fine-grained multi-class classification in accordance with the applicable rules.

Despite the severity of this problem, evasive content detection remains substantially underexplored. The limitations of current models can be attributed to two key factors. First, LLMs and VLMs continue to produce hallucinations~\citep{Huang_2025,bai2025hallucinationmultimodallargelanguage}, fail to follow instructions reliably~\citep{lou2024largelanguagemodelinstruction,murthy2025evaluatinginstructionfollowingabilitieslanguage}, and struggle to process long or multimodal contexts~\citep{wang2024limitssurveytechniquesextend,chen2024longvilascalinglongcontextvisual}, all of which impair the recognition of subtle deceptive content. Second, real-world e-commerce regulations evolve rapidly, and human annotations are frequently inconsistent, resulting in training data that is both noisy and ambiguous. These two factors together hinder the development and evaluation of models suited to reliable content moderation.

To address this gap, we introduce the \textbf{Eva}sive Content \textbf{D}etection in \textbf{E}-Commerce \textbf{Bench}mark (\textbf{EVADE-Bench}), a Chinese multimodal benchmark designed to evaluate the ability of LLMs and VLMs to detect evasive content in real-world e-commerce scenarios. Every sample is iteratively annotated by domain experts to ensure accurate and consistent ground truth labels.

EVADE-Bench provides two evaluation tracks that assess complementary aspects of model competence: the ability to efficiently identify specific policy violations, and robustness when reasoning under complex, multi-rule policy contexts. We conduct a comprehensive evaluation of 26 mainstream LLMs and VLMs on EVADE-Bench and analyze their performance on this challenging task.

\textbf{Our key contributions are as follows:}
\begin{enumerate}
    \item We release EVADE-Bench, the first expert-curated Chinese multimodal dataset tailored for evasive content detection in e-commerce. The benchmark comprises two evaluation tracks, Single-Violation and All-in-One, which probe distinct reasoning capabilities under varying policy contexts.

    \item We demonstrate that clearer rule categorization significantly improves model prediction consistency and reduces false predictions, highlighting the critical role of benchmark design in enabling reliable evaluation.

    \item We establish systematic baselines by evaluating 26 open- and closed-source LLMs and VLMs, providing the first comprehensive reference for this underexplored yet high-impact problem. We further analyze the common error patterns of these models on EVADE-Bench, examine potential limitations of multimodal large models in this setting, and investigate the feasibility of multi-agent decomposition as a strategy for improving reasoning accuracy.
    
\end{enumerate}

Through these contributions, we aim to advance research on adversarial and evasive content detection, support the development of safer and more trustworthy content moderation systems, and promote robust multimodal reasoning in high-stakes commercial applications.

\section{Related Work}

Recent work on online safety highlights adversarial and obfuscated content detection in hate speech and cyberbullying. SWE2~\citep{Mou_2020} enhances robustness to lexical attacks by combining word- and subword-level features, while an LSTM-based model with correction mechanisms improves resilience to deceptive cyberbullying patterns~\citep{azumah2024deeplearningapproachesdetecting}. Autoregressive models have also been used to craft graded adversarial examples for implicit hate detection~\citep{ocampo-etal-2023-playing}, introducing a ``build-it, break-it, fix-it" retraining loop that boosts model robustness to nuanced, context-sensitive abuse.

The safety and robustness of VLMs have gained attention with specialized benchmarks. The Hateful Memes challenge~\citep{kiela2021hatefulmemeschallengedetecting} pioneered rigorous multimodal evaluation by using subtle hateful content contrasted with benign distractors, discouraging unimodal shortcuts.
Later benchmarks like MM-SafetyBench~\citep{liu2024mmsafetybenchbenchmarksafetyevaluation} used 5,040 adversarial image-text pairs to show that even aligned models are vulnerable to malicious prompts. MMSafeAware~\citep{wang2025cantforesttreesbenchmarking} found GPT-4V misclassified over one-third of unsafe and over half of safe inputs, exposing poor safety awareness across 29 threats. VLDBench~\citep{raza2025vldbenchvisionlanguagemodels} evaluated 31,000 news-image pairs and showed that adding visual context improves disinformation detection by 5–35\% and enhances compliance monitoring.

\section{The EVADE-Bench}

We introduce EVADE-Bench, a multimodal benchmark grounded in Chinese advertising law regulations, designed to evaluate whether current LLMs and VLMs can effectively identify evasive content in real-world e-commerce settings. EVADE-Bench integrates both textual and visual inputs, requiring models to jointly comprehend and reason over multimodal data in accordance with explicit policy guidelines.

\begin{table}[!thp]\centering
\small
\caption{Data distribution of EVADE-Bench and the number of rule prompt tokens for each violation category.}
\begin{tabular}{lccc}
\toprule[1.5pt]
\textbf{Category} &\textbf{Text Count} &\textbf{Image Count} &\textbf{Rule Tokens}\\\midrule
Body Shaping &202 &2,134 &614\\
Women's Health &211 &1,295  & 652\\
Height Growth &553  &3,424  &953\\
Men's Health &652  &1,738  &1,123\\
Weight Loss &442  &1,203  &1,364\\
Health Supplement &773 &4,167  &3,379\\\midrule
Overall &2,833 &13,961  &/\\
\bottomrule[1.5pt]
\end{tabular}
\label{tab:dataset}
\end{table}

\subsection{Overview of the EVADE-Bench}

As summarized in \textbf{Table \ref{tab:dataset}}, this benchmark comprises 2,833 text samples and 13,961 images, all collected from real-world e-commerce platforms. Each instance has been manually annotated by domain experts with deep familiarity in advertising law, ensuring high-quality and regulation-compliant labels.

For each sample (image or text) in EVADE-Bench, we provide a corresponding text prompt that contains all rules of the current data type. We construct a multimodal input pair by concatenating each image with its prompt for VLMs' reasoning, and construct a pure text input pair by concatenating each text with its prompt for LLMs' reasoning.

Images present a particularly challenging modality, as they often embed both textual claims and visual cues. Thus, image-based reasoning not only requires the VLMs to correctly extract embedded text via OCR but also to interpret visual elements in conjunction with text prompt. This dual-modality reasoning is essential for detecting subtle forms of evasion, such as cropped disclaimers or euphemistic imagery.
At the same time, texts present unique challenges for LLMs due to intentionally created variations such as missing keywords, homophones, and typos.

\subsection{Data Collection and Rule Formulation}

To construct a diverse and regulation-aligned benchmark, we collected 25,380 raw texts and 48,000 raw images from six e-commerce sub-domains (e.g., body shaping, height growth). Through collaboration with advertising experts, we designed six rules based on Chinese advertising law.
While expert human annotation provides high-fidelity labels, it is not immune to inconsistencies due to fatigue, subjectivity, or ambiguous cases. To ensure dataset uniqueness, diversity, and quality, we introduced a multi-stage pipeline for collecting challenging samples and generating ground-truth from raw images and texts.

\begin{algorithm}[!thp]\centering
\small
\setstretch{1}
\begin{algorithmic}[1]
    \STATE \textbf{Input:} Raw Dataset $D$ without ground truth
    \STATE \textbf{Output:} Annotated Dataset $D_{final}$ with ground truth $GT$
    
    \STATE \textbf{Stage 1: Data Preprocessing}
        \STATE $D_{img} \leftarrow \text{UniqueByID}(D_{images})$
        \STATE $D_{txt} \leftarrow \text{UniqueByID}(D_{texts})$
    
    \STATE \textbf{Stage 2: Clustering \& Sampling}
        \FOR{$modality \in \{img, txt\}$}
            \STATE $E_{modality} \leftarrow \text{CLIP\_Encode}(D_{modality})$
            \STATE $C \leftarrow \text{KMeans}(E_{modality}, K=300)$
            \STATE $D_{balanced}[modality] \leftarrow \bigcup_{c \in C} \text{RandomSample}(c, \min(60, |c|))$
        \ENDFOR
    
    \STATE \textbf{Stage 3: Model Validation}
        \STATE $D_{unlabeled} \leftarrow \{s \in D_{balanced} : \text{HasDisagreement}(s)\}$
    
    \STATE \textbf{Stage 4: Expert Annotation}
        \STATE $D_{final} \leftarrow \emptyset$
        \FOR{each sample $i \in D_{unlabeled}$}
            \STATE $P^i \leftarrow$ Model predictions for sample $i$
            \FOR{attempt $\in \{1, 2, ensemble\}$}
                \STATE $A^i \leftarrow$ Get annotation (initial/revised/5-expert)
                \IF{$A^i$ matches all in $P^i$ OR attempt = ensemble}
                    \STATE $GT^i \leftarrow A^i$ (or mode for ensemble)
                    \STATE \textbf{break}
                \ENDIF
            \ENDFOR
            \STATE Add $(i, GT^i)$ to $D_{final}$
        \ENDFOR
    
    \STATE \textbf{return} $D_{final}$
\end{algorithmic}
\caption{Integrated Dataset Filtering and Annotation Pipeline}
\label{alg:integrated_pipeline}
\end{algorithm}

First, we employed LLMs and VLMs of various sizes to automatically generate a high-quality dataset from raw images and texts.
A sample is considered simple and excluded from $D_{unlabeled}$ if all models, including the smallest one, produce consistent predictions, highlighting the challenging nature of $D_{unlabeled}$.
At this stage, the dataset temporarily lacks ground-truth labels.
Next, trained expert annotators labeled each instance and compared their annotations with predictions from the three best LLMs (GPT-o1mini, DeepSeek-R1, QwenMax) and VLMs (GPT-4o, Claude-3.7, Gemini-2.5-Pro) to identify inconsistent cases. Through iterative comparison and annotation, we ultimately generated ground-truth labels for each instance in the dataset.
The detailed pipeline is presented in \textbf{Algorithm \ref{alg:integrated_pipeline}}.
To prevent dataset contamination and maintain unbiased evaluation, we ensure that experts perform their annotations without access to model predictions.

\section{Experiment}
\subsection{Setting}

\paragraph{Baselines}

We conducted a Human Baseline Experiment with three independent domain experts to annotate all texts and images from EVADE-Bench and calculated Krippendorff's Alpha to analyze human annotation quality.
Due to its 4K context length limitation, DeepSeek-VL2-27B was only used for the Single-Violation task.

\paragraph{Tasks} 
The evaluation is structured around two core tasks: Single-Violation and All-in-One, designed to probe distinct capabilities in complex reasoning.

\paragraph{Single-Violation Task}

The task evaluates model performance across six distinct product categories using short, domain-specific prompts (e.g., 202 texts and 2,134 images assessed with a 614-token prompt for body shaping products). It tests the model's fine-grained reasoning ability within narrowly defined contexts.
However, the task faces a key challenge: semantic overlaps between categories (e.g., ``weight loss" products often claim health benefits, while ``health improvement" products frequently emphasize weight management as a health outcome) can confuse the model's decision-making. To simulate more complex, classification-dense scenarios, we propose the All-in-One task.

\paragraph{All-in-One Task}

We unify prompts across six violation types into a single instruction, expanding prompt length from 1K to 7K tokens and increasing classification labels from an average of 5 (in Single-Violation) to 26 distinct regulatory categories.
We merge semantically overlapping rules to reduce ambiguity. This means increasing both the context length and the number of classifications during model inference, but reducing the possibility of confusion.
To assess model generalization in adversarial e-commerce scenarios, we define two sub-tasks: the Simplified Instruction task and the Detailed Instruction task.
\textbf{Simplified Instruction} refers to the approach where, in the input prompt, we avoid introducing any examples except for the necessary definition. 
The purpose of this approach is to allow models for free reasoning based on the prompt, in order to explore the upper and lower bounds of its performance.
\textbf{Detailed Instruction} refers to the approach where, in the input prompt, we not only include the necessary definition but also introduce positive and negative examples.
The purpose of this approach is to constrain the model’s free-form generation through detailed examples and descriptions, thereby stabilizing its performance.

\subsection{Evaluation indicators}

Since our task requires models to analyze EVADE-Bench samples and provide final classifications, we use two metrics: Full Accuracy and Partial Accuracy.
Where $N$ is the total number of samples, $C_i$ is the predicted set, $G_i$ is the ground truth set, and $\mathbb{I}(\cdot)$ is an indicator function.

\begin{align}
  \text{Acc}_{full} &= \frac{1}{N} \sum_{i=1}^{N} \mathbb{I}(C_i = G_i) \label{eq:full_accuracy} \\
  \text{Acc}_{partial} &= \frac{1}{N} \sum_{i=1}^{N} \mathbb{I}(C_i \cap G_i \neq \varnothing) \label{eq:partial_accuracy}
\end{align}

In e-commerce, these metrics serve different purposes. Full Accuracy measures model-human alignment, while Partial Accuracy addresses the practical challenge that categories lack clear boundaries. When dealing with metaphorical content, semantic overlaps are inevitable, and pursuing extreme Full Accuracy may cause overfitting. In practice, moderators focus on identifying rule violations rather than specific violation types.

Therefore, while Full Accuracy aligns with mainstream benchmarks, Partial Accuracy better reflects practical e-commerce applications.

\begin{table}[!thp]\centering
\small
\caption{All model overall performance on Single-Violation of EVADE-Bench.}
\setlength{\tabcolsep}{5pt} 
\begin{tabular}{lc|lc}
\toprule[1.5pt]
\textbf{LLMs} & \textbf{Part. / Full.} &\textbf{VLMs} & \textbf{Part. / Full.}\\
\midrule
\multicolumn{4}{c}{\textit{Open Source}}\\\midrule

Llama-3.1-8B & 35.62 / 20.93 & MiniCPM-V2.6 &44.15 / 12.37\\
LLama-3.1-70B &38.55 / 26.23 & InternVL3-8B & 42.48 / 18.87 \\
Qwen2.5-7B & 38.48 / 27.18 & InternVL3-14B & 51.20 / 23.23\\
Qwen2.5-14B &45.39 / 28.70 & InternVL3-38B & 49.19 / 21.40 \\
Qwen2.5-32B & 46.56 / \textbf{29.69} & DeepSeek-VL2 & 29.12 / 12.24\\
Qwen2.5-72B & 49.21 / 27.85 & Qwen2.5-VL-7B & 44.52 / 19.96\\
DeepSeek-V3 & 51.85 / 28.77 & Qwen2.5-VL-32B & 52.39 / 22.57 \\
DeepSeek-R1 & \textbf{54.64} / 25.45 & Qwen-VL-72B & \textbf{57.63} / \textbf{25.05}\\
\midrule
\multicolumn{4}{c}{\textit{Close Source}}\\
\midrule
GPT-o1mini & 49.28 / 29.72 & GPT-4o & 58.47 / \textbf{26.96}\\
GPT-4.1 & \textbf{52.74} / \textbf{31.59} & Claude-3.7 & \textbf{58.79} / 23.42 \\
Qwen2.5-max & 48.29 / 31.27 & Qwen2.5-VL-max & 53.38 / 25.60\\
 & &Gemini-2.5-pro & 52.44 / 22.14\\\midrule
 
\multicolumn{4}{c}{\textit{Human Performance}}\\\midrule
Human on Text &69.34 / 57.03 &Human on Image &69.20 / 51.41\\\midrule
\multicolumn{4}{c}{\textit{Human Annotation Krippendorff's Alpha Scores}}\\\midrule
Human on Text &0.6867 &Human on Image &0.7513\\

\bottomrule[1.5pt]
\end{tabular}
\label{tab:singleriskoverall}
\end{table}

\section{Main Results}

\subsection{Single-Violation Results}

For VLMs processing image-text pair inputs, as shown in \textbf{Table~\ref{tab:singleriskoverall}}, the closed-source models Claude-3.7-sonnet and GPT-4o achieve the highest overall accuracy across all six categories. Among open-source VLMs, Qwen2.5-VL-72B demonstrates the best performance, showing competitive results despite its open-source nature. In contrast, DeepSeek-VL2-27B shows the weakest performance, significantly lagging behind even smaller models such as 8B-parameter VLMs, highlighting the impact of model architecture and training strategies rather than model size alone.

For LLMs processing pure text inputs, the LLaMA series struggles significantly due to limited Chinese language capabilities. Even the 70B version of LLaMA performs only at par with the 7B version of the Qwen series, revealing a critical limitation in multilingual robustness for otherwise powerful models.

Qwen3\citep{yang2025qwen3technicalreport} is a released series of LLMs that integrates both \textit{thinking} and \textit{non-thinking} modes within a single architecture. However, as shown in \textbf{Table~\ref{tab:qwen3_singlerisk_result}}, its largest variant, Qwen3-235B-A22B, still underperforms compared to DeepSeek-R1-671B and DeepSeek-V3-671B. Surprisingly, enabling the thinking mode yields minimal improvements and even leads to performance degradation in some models, such as Qwen3-32B.

\begin{table}[!thp]\centering
\small
\setlength{\tabcolsep}{6pt} 
\caption{
Qwen3 models' performance on EVADE-Bench Single-Violation.}
\begin{tabular}{lccc}
                
\toprule[1.5pt]
\textbf{Model} &\textbf{Mode}  &\textbf{Partial Accuracy} &\textbf{Full Accuracy}\\
\midrule
Qwen3-32B &Flat &48.71 & 26.16\\
Qwen3-32B &Think & 47.86& 24.71 \\\midrule
Qwen3-30B-A3B &Flat &46.70 & 26.44\\
Qwen3-30B-A3B &Think & 47.51 & 27.64 \\\midrule
Qwen3-235B-A22B &Flag & 49.77 &27.46 \\
Qwen3-235B-A22B &Think & 50.02 & 27.50 \\
\bottomrule[1.5pt]
\end{tabular}
\label{tab:qwen3_singlerisk_result}
\end{table}

The Krippendorff's Alpha scores (0.67 $\sim$ 0.8) indicate substantial agreement among experts, particularly on images (at least 0.75 overall), proving that despite the task's complexity, the annotation rules are well-defined and consistent for humans, which dispels concerns about label noise. Although models perform competitively on certain categories, humans still outperform all LLMs and VLMs overall in understanding evasive text and images (Partial Acc.) and also surpass current models in precision (Full Acc.).

A critical observation across all models is the significant gap between partial accuracy and full accuracy, often exceeding 10\%.
This gap shows how models struggle to achieve complete understanding: while they can catch basic meanings or simple features, they often miss important details in meaning or images, especially when dealing with indirect language, hidden meanings, or misleading images.
While closed-source models generally perform better than open-source ones, there are clear exceptions. For instance, DeepSeek-R1 shows strong results in text understanding, even performing better than some closed-source models, while Qwen2.5-VL-72B shows good ability in image understanding, though not quite reaching the level of GPT-4o or Claude-3.7-sonnet.

\subsection{All-in-One Results}

We observe substantial performance improvements across models, particularly in smaller LLMs and VLMs, as shown in \textbf{Table~\ref{tab:allinone_result}}. Without altering the input data, merely combining overlapping classifications significantly enhances the models' reasoning capabilities. Despite a sixfold increase in prompt length (compared to the Single-Violation setting) and an expansion from a few classifications to 26, models generally demonstrate improved rather than degraded performance.
This suggests that the clarity of boundaries between classifications poses a more significant challenge for models than the context length or the number of classifications.
Notably, the performance gap among LLMs narrows significantly in the All-in-One setting. Additionally, the difference between partial and full accuracy decreases dramatically, dropping from over 10\% in the Single-Violation setting to approximately 5\% in All-in-One.

\begin{table}[!thp] 
\centering
\small
\caption{
The performance of all models on the All-in-One task of EVADE-Bench.
``Simp. Instr." represents Simplified Instruction, while ``Det. Instr." represents Detailed Instruction.
}
\scalebox{0.81}{
\setlength{\tabcolsep}{2pt} 

\begin{tabular}{lcclcc}
\toprule[1.5pt]
\multirow{2}{*}{\textbf{LLMs}} & \multicolumn{2}{c}{\textbf{(Part / Full) Accuracy}} & \multirow{2}{*}{\textbf{VLMs}} & \multicolumn{2}{c}{\textbf{(Part / Full) Accuracy}} \\
\cmidrule{2-3} \cmidrule{5-6}
& \textbf{Simp. Instr.} & \textbf{Det. Instr.} & & \textbf{Simp. Instr.} & \textbf{Det. Instr.} \\\midrule
\multicolumn{6}{c}{\textit{Open Source}}\\\midrule

Llama3.1-8B & 41.94 / 36.47 & 37.84 / 34.44  & MiniCPM-V2.6 & 46.70 / 43.22 & 39.60 / 36.93 \\
LLama3.1-70B & 47.81 / 44.53 & 47.00 / 44.60 & InternVL3-8B & 56.26 / 53.90 & 56.97 / 54.17 \\
Qwen2.5-7B & 51.68 / 49.52 & 53.18 / 50.60 & InternVL3-14B & 60.63 / 56.83 & 60.67 / 57.41 \\
Qwen2.5-14B & 53.14 / 48.32 & 55.23 / 51.20 & InternVL3-38B & 60.62 / 57.35 & 61.36 / 58.34 \\
Qwen2.5-32B & 55.12 / \textbf{51.20} & 53.55 / 49.95 & Qwen2.5VL-7B & 53.15 / 51.49 & 53.39 / 51.20 \\
Qwen2.5-72B & 56.93 / 51.06 & 55.59 / 50.05 & Qwen2.5VL-32B & 62.04 / 58.55 & 61.73 / 59.17 \\
DeepSeek-V3 & 56.58 / 50.62 & \textbf{58.79 / 52.65} & Qwen2.5VL-72B & \textbf{64.25 / 59.09} & \textbf{63.85 / 59.44} \\
DeepSeek-R1 & \textbf{58.25} / 49.35 & 58.69 / 50.37 &  &  & \\\midrule
\multicolumn{6}{c}{\textit{Close Source}}\\\midrule

GPT-o1mini & 56.69 / 51.15 & 54.17 / 48.76 & GPT4o-0806 & 64.14 / 58.05 & 65.12 / \textbf{59.58} \\
GPT4.1-0414 & \textbf{59.16} / 53.79 & \textbf{59.64} / \textbf{53.67} & Claude-3.7 & 64.58 / 56.83 & 63.05 / 56.05 \\
Qwen-max & 58.35 / \textbf{54.68} & 56.48 / 52.88 & QwenVL-max & 63.58 / \textbf{59.24} & 63.31 / 59.50 \\
 &   &  & Gemini2.5-pro & \textbf{70.57} / 54.45 & \textbf{70.43} / 51.94 \\

\bottomrule[1.5pt]
\end{tabular}
}
\label{tab:allinone_result}
\end{table}

Interestingly, while high-performing LLMs—such as DeepSeek-R1, DeepSeek-V3, and the GPT series—can achieve notable improvements in full accuracy, their gains in partial accuracy under the All-in-One are relatively modest.

This may be attributed to ceiling effects or performance saturation that limit further improvements. In contrast, previously underperforming models show improvements across both metrics. For example, Qwen2.5-7B improves its partial accuracy from 38.48\% to 53.18\% and full accuracy from 27.18\% to 50.60\%, achieving over 10\% gains in both metrics. Similarly, Llama-3.1-70B, despite its limited proficiency in Chinese, improves from 38.55\% to 47.00\% in partial accuracy and from 26.23\% to 44.60\% in full accuracy.

This trend also extends to VLMs: previously underperforming models like InternVL3-8B and MiniCPM-V2.6-8B demonstrate dramatic improvements, while large-scale models continue to achieve further gains. These results affirm the effectiveness of the All-in-One setting in reducing confusion caused by overlapping classifications and reveal the potential of structured prompt engineering for regulatory tasks.

\subsection{Analysis of the Effect of RAG and SFT}

Given that EVADE-Bench contains fewer text than image samples, we investigate model reasoning enhancement through Retriever-Augmented Generation (RAG) for text and Supervised Fine-tuning (SFT) for images. We conducted RAG on LLMs and SFT exclusively on Qwen-VL series for VLMs.
We split text samples into document and query sets (2:8 ratio), then identified the most similar document for each query using text similarity to enhance LLM reasoning. For SFT, we split image samples into training (80\%) and test (20\%) sets, fine-tuning VLMs for three epochs.

\begin{figure}[!htb]\centering
\centering
\includegraphics[width=\linewidth]{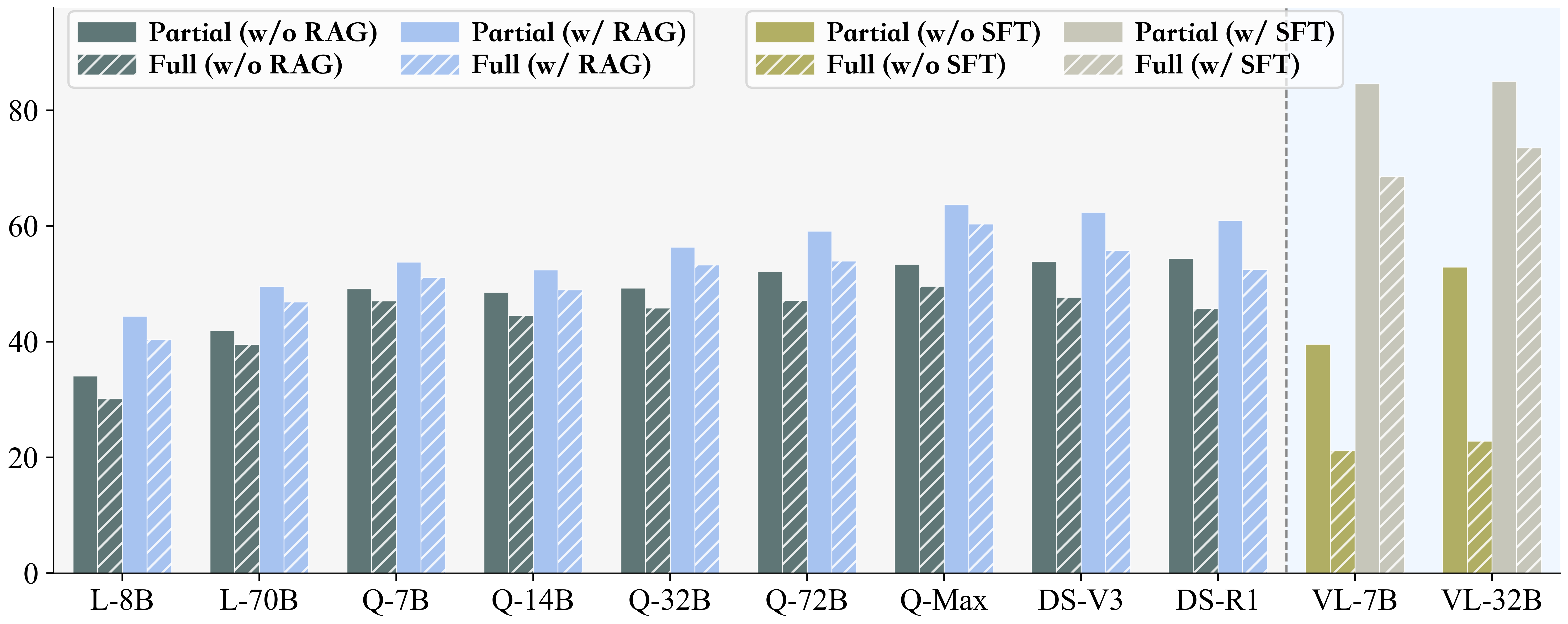}
\caption{Comparison of LLMs and VLMs before and after the introduction of RAG and SFT. The models compared include: Llama-3.1 (abbreviated as L), Qwen2.5 (Q), DeepSeek (DS), and Qwen2.5-VL (VL).}
\label{fig:rag_and_sft}
\end{figure}

As shown in \textbf{Figure~\ref{fig:rag_and_sft}}, both RAG and SFT significantly improved model performance on Single-Violation and All-in-One tasks. RAG and SFT enhance model precision by providing semantically aligned reference materials, especially for ambiguous or metaphorical inputs.

\subsection{Error Analysis}
\label{erroranalysis}

Analyzing Qwen-2.5-VL-7B's performance in \textbf{Table~\ref{tab:error_analysis}}, we find VLMs struggle with missing key information or indirect implications. VLMs must identify text through OCR before inferring missing information, but image distortion disrupts this process. Models show insufficient sensitivity to metaphors, euphemisms, and dual meanings, and poorly recognize embedded text obscured by noise or masking. Comparing pre- and post-SFT performance, SFT significantly reduces these errors by providing additional knowledge about violating images.

\begin{table}[!thp]\centering

\caption{
Randomly select 100 error types and categorize by cause: model hallucinations on non-existent elements (Hallu.); failure to locate key information (Focus.); intentional information omissions/errors (Omit.); recognition failures from occlusion/deformation (Distort.); correct detection but failure to find prompt rules (Find.); and correct detection but rule misunderstanding (Match.).
}
\scalebox{0.9}{
\begin{tabular}{lcccccc}
\toprule[1.5pt]
\textbf{Model} &\textbf{Hallu.} &\textbf{Focus.} &\textbf{Omit.} &\textbf{Distort.} &\textbf{Find.} &\textbf{Match.}\\\midrule
QVL-7B &1 &4 &26 &5 &38 &26\\
QVL-7B$_{\scriptstyle\mathrm{sft}}$  &0 &0 &8 &0 &6 &3 \\\midrule
QVL-72B &0 &0 &26 &6 &36 &32\\
GPT-4o &1 &3 &13 &4 &46 &33\\
\bottomrule[1.5pt]
\end{tabular}
}
\label{tab:error_analysis}
\end{table}

We observe that stronger models more frequently misinterpret prompt rules. GPT-4 identifies key violating words but fails to match them to classification criteria. This reveals two challenges: (1) models struggle with backward reasoning, generating information but failing to match it back to prompt requirements; (2) models show systematic biases in prompt interpretation versus human understanding. Stronger models' higher confidence leads them to adhere more firmly to conclusions, amplifying deviations from human understanding and increasing errors.

Although EVADE-Bench is multi-classification, it includes a critical implicit rule: Z.other (no violation) cannot co-exist with any violation classification. Models must understand this constraint before classification. Weaker models violate this rule, incorrectly labeling items as both violation and non-violation, revealing limited ability to understand complex classification constraints.

\subsection{Decomposing Multimodal Reasoning}

We have previously analyzed the reasoning failures in current multimodal large models. We hypothesize that the fundamental cause of these errors lies in the training process of VLMs. While building upon LLMs has enabled new multimodal reasoning capabilities, it has potentially weakened the text processing abilities - a perspective validated by numerous studies\citep{guo2025integratingvisualinterpretationlinguistic, zheng2025mllmsdeeplyaffectedmodality}. Prism\citep{qiao2024prismframeworkdecouplingassessing} proposed decomposing multimodal reasoning into two stages: ``VLM generating textual descriptions of image information" followed by ``LLM performing reasoning on the text." This approach significantly improved model performance in multimodal reasoning tasks.

\begin{figure}[!htb]\centering
\centering
\includegraphics[width=\linewidth]{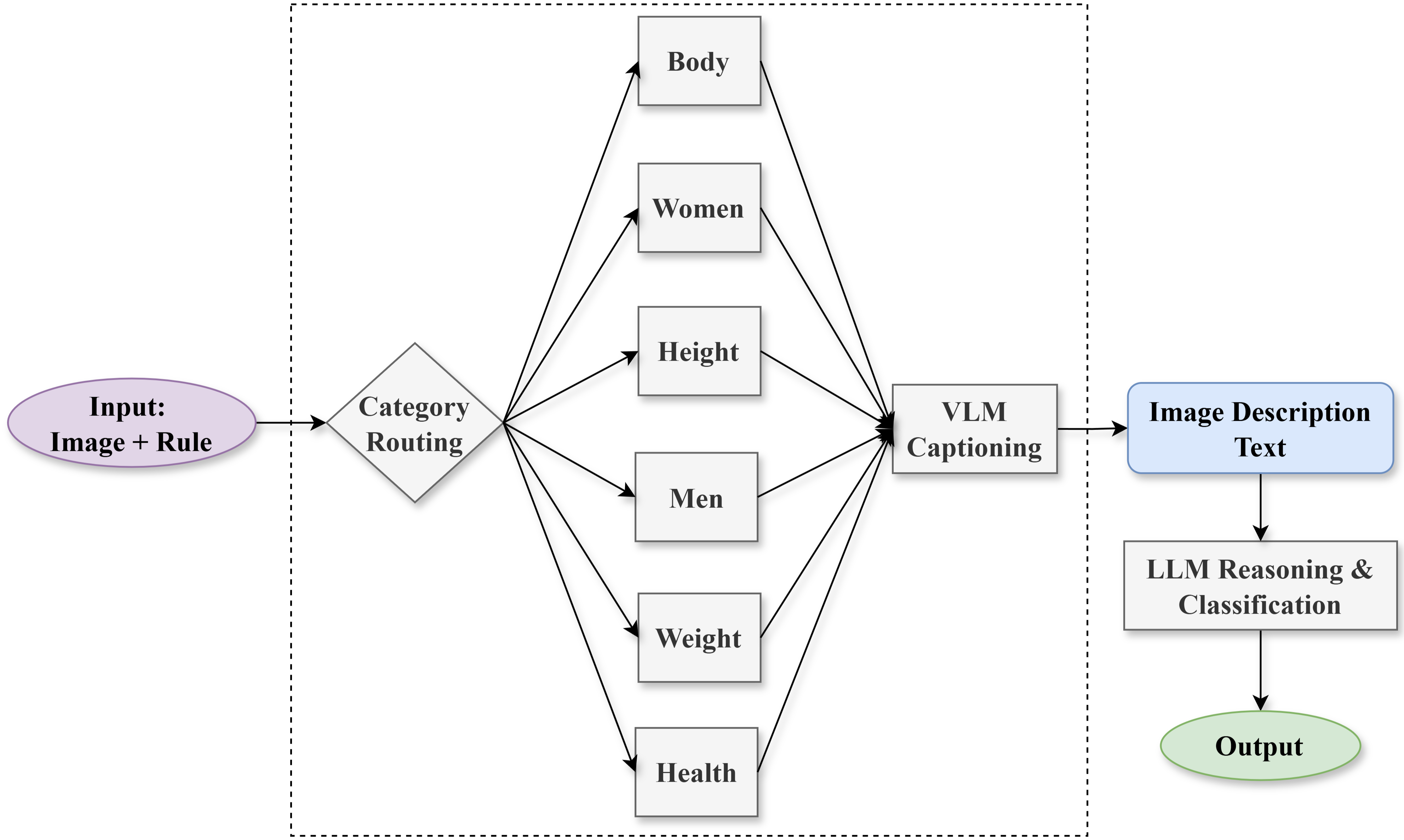}
\caption{Multimodal decomposition and inference process.}
\label{fig:vlm_decomposing}
\end{figure}

However, it remains unknown whether this method is still effective on metaphorical images like EVADE-Bench, which are obviously lacking contextual information.

To verify the effectiveness of this approach, we conducted experiments using GPT-4o and Claude-3.7 specifically for converting images into pure text descriptions, and then employed different powerful LLMs to perform multi-class inference on the text generated by GPT-4o and Claude-3.7 alone, without access to the images themselves. As \textbf{Table \ref{tab:decompose_detail}} shows, compared to single-stage multimodal reasoning, using the same VLM to describe images and then passing the descriptions to a dedicated LLM for reasoning leads to better performance.

\begin{table}[!thp]\centering
\setlength{\tabcolsep}{10pt} 
\renewcommand{\arraystretch}{1} 
\small
\caption{Decomposing multimodal reasoning performance comparison}
\begin{tabular}{lcc}
\toprule[1.5pt]
\textbf{Model} & \textbf{Partial Accuracy} & \textbf{Full Accuracy} \\\midrule
\multicolumn{3}{c}{\textit{Multimodal reasoning without decomposition}}\\\midrule
GPT4o & 59.61 & 29.14  \\
Claude-3.7 & 59.07 & 24.53  \\\midrule
\multicolumn{3}{c}{\textit{GPT-4o as the image description model}}\\\midrule
GPT4.1 & \textbf{61.65} & \textbf{32.76}  \\
Qwen3-235b & 61.30 & 28.64   \\
DeepSeek-R1 & 60.87 & 28.61  \\
\midrule
\multicolumn{3}{c}{\textit{Claude-3.7 as image description model}}\\\midrule
GPT4.1 & \textbf{62.55} & \textbf{30.97}  \\
Qwen3-235b & 61.19 & 24.96  \\
DeepSeek-R1 & 59.86 & 24.74  \\
\toprule[1.5pt]
\end{tabular}
\label{tab:decompose_detail}
\end{table}

\section{Conclusion}

EVADE-Bench is the first Chinese-language multimodal benchmark for detecting evasive e-commerce content, covering expert-annotated texts and images across six categories. Evaluation of 26 LLMs and VLMs reveals significant performance disparities and shows that both LLMs and VLMs still fall short on EVADE-Bench. The introduction of EVADE-Bench is therefore poised to catalyze progress in this critical area.
We find that merging overlapping classifications enhances reasoning and reduces the accuracy gap, despite expanding prompt length and increasing classification labels—particularly in smaller models, underscoring the importance of rule clarity.
Our analysis reveals common failure modes in EVADE-Bench, including contextual noise, obfuscated language, and OCR errors, indicating weaknesses in semantic understanding and vision-language alignment, and explored the feasibility of decomposing ``single multimodal inference" into ``description before inference" through Multi Agent.
Looking forward, EVADE-Bench offers a rigorous foundation for evaluating and advancing multimodal moderation systems, highlighting key bottlenecks and design principles—such as clearer taxonomy and robust error handling—for building safer, more reliable AI systems.

\section*{Acknowledgments}

Min Yang was supported by the Guangdong S\&T Program (No. 2025B0101130005), the Natural Science Foundation of Guangdong Province of China (2024A1515030166, 2025B1515020032), and the Innovation Team Project of Guangdong Province (No. 2024KCXTD017).
This work was supported by CCF-Alimama Tech Bag Fund.

\bibliographystyle{ACM-Reference-Format}

\balance  

\bibliography{sample-base}










\end{document}